\documentclass[]{kling/kling}

\usepackage{microtype}
\usepackage{graphicx}
\usepackage{subcaption}
\usepackage{csquotes}
\usepackage{afterpage}
\usepackage{xcolor}

\usepackage{amsmath}
\usepackage{amssymb}
\usepackage{mathtools}
\usepackage{amsthm}
\usepackage{xspace}
\usepackage{colortbl}
\usepackage{multirow}
\usepackage{multicol}
\usepackage{enumitem}
\usepackage{wrapfig}
\usepackage{nicefrac}
\usepackage[font=small,labelfont=bf,justification=centering]{caption}

\usepackage{booktabs}
\usepackage{multirow}
\usepackage{graphicx}
\usepackage{xcolor}
\usepackage{pifont} 

\usepackage{geometry}
\usepackage{tabularx} 

\usepackage{amsmath}  

\title{KlingAvatar 2.0 Technical Report}
\headertext{KlingAvatar 2.0: Co-Reasoning Directed Spatio-Temporal Cascade Modeling for Avatar Synthesis}

\author[]{Kling Team, Kuaishou Technology}

\abstract{
Avatar video generation models have achieved remarkable progress in recent years. However, prior work exhibits limited efficiency in generating long duration high-resolution videos, suffering from temporal drifting, quality degradation, and a weak prompt follow-up as the video length increases. To address these challenges, we propose \textbf{KlingAvatar 2.0}, a spatio-temporal cascade framework that performs upscaling in both spatial resolution and temporal dimension by first generating low-resolution blueprint video keyframes that capture global semantics and motion, and then refining them into high-resolution, temporally coherent sub-clips using a first-last frame strategy, while retaining smooth temporal transitions in long-form videos. To enhance cross-modal instruction fusion and alignment in extended long videos, we introduce a Co-Reasoning Director composed of three modality-specific large language model (LLM) experts. These experts reason about modality priorities and infer the underlying user intent, converting inputs into detailed storylines through multi-turn dialogue. A negative director further refines negative prompts to improve instruction alignment. Building on these components, we extend the framework to support ID-specific multi-character control. Extensive experiments demonstrate that our model effectively addresses the challenges of efficient, multimodally aligned long-form high-resolution video generation, delivering enhanced visual clarity, realistic lip–teeth rendering with accurate lip synchronization, strong identity preservation, and coherent multimodal instruction following.
}

\metadata[Date]{December 15, 2025}
\metadata[Access]{\url{https://app.klingai.com/global/ai-human/image/new/}}
\metadata[Model ID]{Avatar 2.0}

\begin{document}

\maketitle

\section{Introduction}
\label{section:intro}

Audio-driven avatar video synthesis aims to generate realistic and expressive human-centric videos, featuring synchronized facial and emotion expressions, coherent lip-teeth movements and body gestures, and natural character interactions with both the environment and other characters. This technology holds significant value across diverse domains, including education, personalized services, industrial training, entertainment, and advertising, where it enables more immersive and engaging experiences.

The field of audio-driven avatar video generation has evolved significantly through several stages of technological advancement. Early approaches focused primarily on lip-synchronized facial animation, leveraging audio-to-motion representations~\cite{he2023gaia,zhang2023sadtalker,guo2024liveportrait,wang2024vexpress,jiang2024mobileportrait,chen2025echomimic} and direct audio-to-video synthesis for portrait animation~\cite{fada,xu2024hallo,xu2024vasa,stypulkowski2024diffused,tian2024emo,jiang2025loopy,peng2025omnisync}. Subsequent developments expanded the scope to semi-body video generation, incorporating hand gestures and upper body animation~\cite{meng2025echomimicv2,lin2025cyberhost,tian2025emo2}, followed by full-body video generation with background environments and character-environment interactions~\cite{hallo3,cui2025hallo4,gan2025omniavatar,wang2025fantasytalking,lin2025omnihuman1,fei2025skyreels,gao2025wans2v,jiang2025omnihuman1.5,ding2025kling-avatar}. These advances in complex human motion modeling, realistic environment generation, and natural interactions have been largely enabled by pretrained DiT-based video diffusion models~\cite{zheng2024open,yang2025cogvideox,luminavideo,ma2025step,wan2025wan}. To address more complex real-world scenarios and improve controllability, recent works have explored human-object interactions~\cite{huang2025hunyuanhoma} and multi-person conversational scenarios with per-character audio control~\cite{kong2025multitalk,wei2025mocha,wang2025interacthuman}. Multimodal large language model (MLLM) driven storyline planning has emerged as a promising direction, enabling fine-grained expressions, vivid emotions, reasoned actions, and environmental interactions through shot-level guidance for long-duration generation~\cite{jiang2025omnihuman1.5,ding2025kling-avatar}. Despite these advances, existing methods remain inefficient for generating long-duration, high-resolution digital human videos, often suffering from visual degradation and limited coherence with complex, long-horizon multimodal instructions.

\begin{figure}[t]
    \centering
    \includegraphics[width=\textwidth]{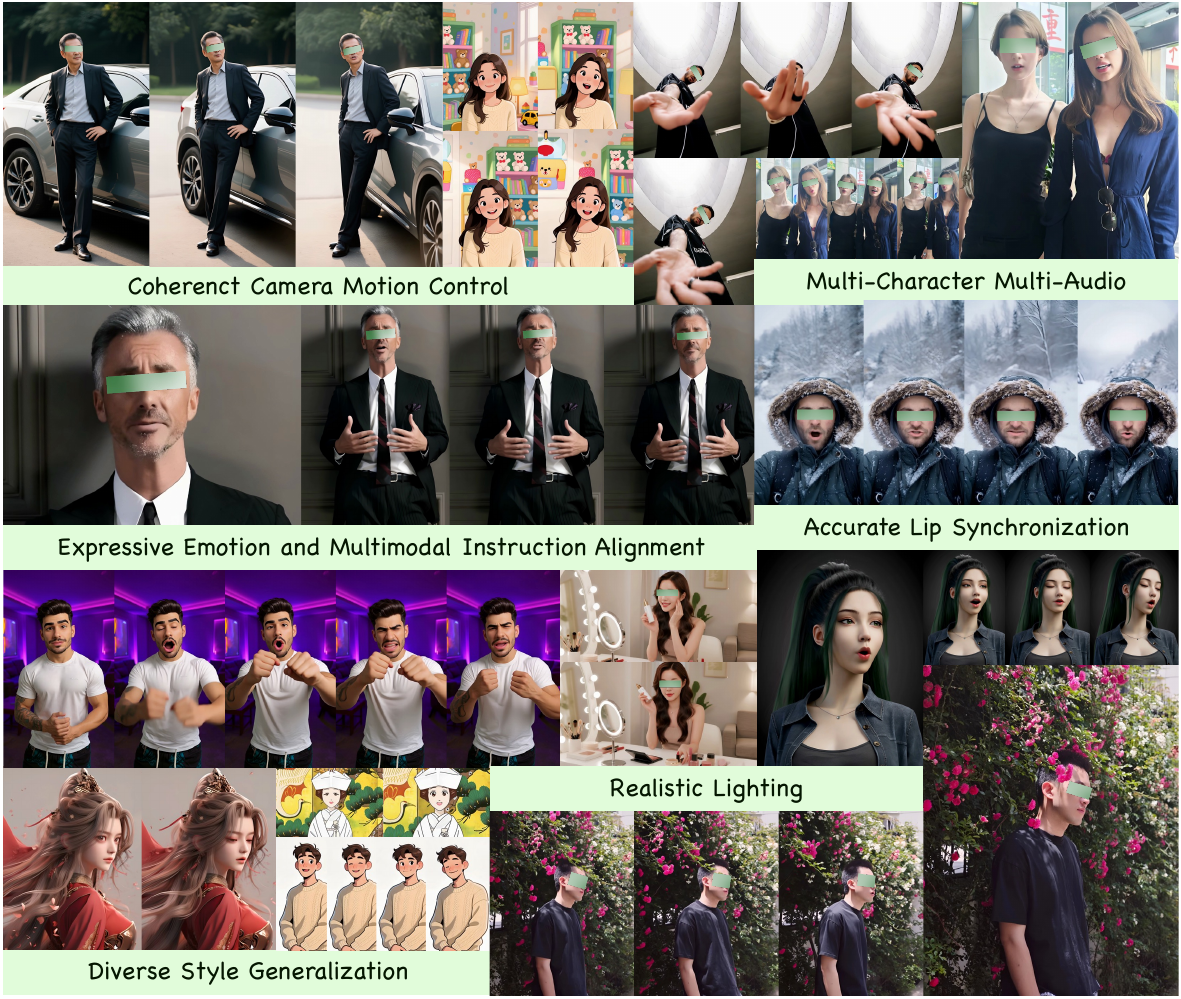}
    \caption{\textbf{KlingAvatar 2.0} generates vivid, identity-preserving digital humans with accurate camera control, expressive emotions, high-quality motion, and precise facial–lip and audio synchronization. It achieves coherent alignment across audio, image, and text instructions, generalizes to diverse open-domain styles, and supports multi-character synthesis with identity-specific audio control. These capabilities are enabled by our multimodal instruction-following, omni-directed spatial–temporal cascade framework for high-resolution, long-duration video generation.}
    \label{fig:teaser}
\end{figure}

Building upon the foundation of~\cite{ding2025kling-avatar}, we propose a unified framework that addresses the aforementioned challenges. To enable efficient long-form, high-resolution video generation, we introduce a temporal and spatial cascade framework that samples low-resolution blueprint videos for efficient generation and then gradually upsamples them to longer durations and higher resolutions. This approach enriches visual details while mitigating temporal drifting artifacts that plague long-form video generation. For long-duration video generation, effective multimodal fusion and adherence to user instructions are crucial. When handled poorly, modality conflicts can cause severe degradation. To address this, we develop a Co-Reasoning Director that improves upon existing MLLM reasoning capabilities, operating in a multi-turn dialogue manner to generate coherent shot-level storylines. This is complemented by a novel negative director that captures fine-grained negative prompts to enhance instruction-following accuracy. Moreover, complex digital human applications often involve multiple characters, where how to accurately drive each human becomes an essential problem. To this end, we leverage deep DiT block features and ID-aware attention to realize mask-controlled audio injection, enabling synchronized yet individually controlled character animations in complex conversational settings. Together, these components form an integrated system that advances the state-of-the-art in audio-driven avatar video synthesis by simultaneously addressing efficiency, instruction alignment, and multi-character coordination.

To develop and train such a model, we curated an enhanced dataset that expands upon~\cite{ding2025kling-avatar}, featuring a substantially larger collection of high-quality, cinematic-level video data. Our dataset encompasses multilingual and multi-character conversational scenarios, with extensive filtering pipelines applied to ensure high visual fidelity and consistent audio-lip synchronization. We conduct extensive experiments to evaluate the performance of our proposed framework. Our evaluation demonstrates that our model achieves superior performance against leading competitors~\cite{jiang2025omnihuman1.5,ding2025kling-avatar,heygen} across visual quality, camera movement, lip synchronization accuracy, fine-grained lip-teeth detail preservation, vivid and natural character animation, and audio-emotion alignment. In terms of generation efficiency, our spatial-temporal cascade framework enables long-duration and high-resolution video synthesis with improved computational efficiency compared to prior methods, while maintaining identity consistency and story continuity for videos up to 5 minutes. We highlight representative generation results in Figure~\ref{fig:teaser}. We summarize our contributions as follows:

\begin{itemize}
\item \textbf{Spatial-Temporal Cascade Framework}: We introduce a spatial-temporal cascade framework that enables efficient generation of long-duration, high-resolution videos through progressive and parallel temporal and spatial upsampling, effectively mitigating temporal drifting artifacts while enriching visual details.
\item \textbf{Co-Reasoning Director}: We develop a Co-Reasoning Director that coordinates multiple MLLMs and LLMs in a multi-turn dialogue manner to capture details across modalities and resolve modality conflicts, generating coherent shot-level storylines. A complementary negative director further enhances fine-grained negative prompts to improve instruction-following accuracy and character emotion expression.
\item \textbf{Multi-Character Multi-Audio Control}: We propose a multi-character, multi-audio control mechanism that exploits deep DiT features for character mask prediction, enabling synchronized yet individually controlled animations from multiple audio streams.
\item \textbf{Strong performance and generalization}: KlingAvatar 2.0 achieves state-of-the-art performance across multiple dimensions, including visual quality, coherent and vivid character animations, natural camera movements, accurate lip synchronization, and precise audio-emotion alignment, while demonstrating strong generalization to diverse domains and scenarios.
\end{itemize}

\section{Related Works}

\noindent\textbf{Video Generation.}
Visual content generation has achieved remarkable breakthroughs through diffusion models, from photorealistic image synthesis to video generation. Early video generation approaches extended pretrained U-Net~\cite{ronneberger2015unet}  based image synthesis models~\cite{ho2020ddpm,song2020denoising,dhariwal2021diffusion,rombach2022high,chen2023pixartalpha,chen2024pixartdelta} by incorporating temporal dimensions or combining 1D temporal attention with 2D spatial attention blocks to capture inter-frame correspondences and reducing computational costs~\cite{guo2023animatediff,wang2023modelscope,ho2022video,singer2022make,blattmann2023stable}. However, such designs that model all frames without temporal compression face limitations in scalability, along with issues of temporal drifting and visual artifacts. Recent DiT-based image synthesis methods~\cite{peebles2023scalable,esser2024scaling,chen2024pixartsigma} have advanced realistic image generation through scalable training paradigms, enabling high-fidelity appearance and improved instruction following. Building upon these progresses, video diffusion models have shifted focus toward DiT-based architectures~\cite{zheng2024open,yang2025cogvideox,luminavideo,ma2025step,wan2025wan}. These methods employ 3D convolutional VAEs to compress videos both temporally and spatially into compact tokens, and leverage large transformer models combined with growing training data and computational resources to capture temporal dynamics and visual details, establishing new state-of-the-art results for video generation. Recent extensions of video diffusion models further advance efficient and long-context generation~\cite{kong2024hunyuanvideo,zhang2025packing,teng2025magi,chen2025midas,zhao2025real,gu2025far}, unified multimodal conditioning with cascaded super-resolution for high-resolution synthesis~\cite{hunyuanvideo2025}, and world modeling~\cite{yu2025context,team2025hunyuanworld,huang2025vid2world}. Despite these advances, these methods are primarily designed for general video generation with text or image prompts, lacking audio conditioning, and remain inadequate for speech-driven digital human modeling.

\noindent\textbf{Multimodal Avatar Synthesis.}
Multimodal avatar synthesis has achieved significant progress through rapid development. The field has evolved from early non-audio driven methods~\cite{siarohin2019firstorder,mraa,zhao2022tps,hu2023animateanyone,facev2v,qiu2025skyreels} that transfer motion from reference videos or landmark sequences, as well as 3D-based digital human synthesis approaches~\cite{chen2025cafe,hu2025ggtalker,cui2025cfsynthesis}, to modern audio-driven video generation systems. Among audio-driven methods, some employ explicit facial landmarks derived from audio features for precise control~\cite{wang2024vexpress,jiang2024mobileportrait,chen2025echomimic}, while others learn implicit motion representations from audio to enable more flexible animation~\cite{he2023gaia,zhang2023sadtalker,guo2024liveportrait}. More recently, transformer-based audio-driven approaches utilize cross-attention mechanisms to eliminate intermediate motion representations, directly generating talking avatars from audio using diffusion models with end-to-end synthesis~\cite{fada,xu2024hallo,xu2024vasa,stypulkowski2024diffused,tian2024emo,jiang2025loopy,peng2025omnisync}. Beyond facial lip synchronization, semi-body methods synchronize hand gestures and upper body movements with audio~\cite{meng2025echomimicv2,lin2025cyberhost,tian2025emo2}, enabling more expressive digital human generation. Recent advances leverage large-scale pretrained video diffusion models to achieve enhanced temporal coherence and visual fidelity~\cite{hallo3,cui2025hallo4,gan2025omniavatar,wang2025fantasytalking,lin2025omnihuman1,fei2025skyreels,gao2025wans2v}, supporting the generation of complex background environments and full body interactions. Further developments include specialized extensions for human-object interactions~\cite{huang2025hunyuanhoma} and multi-person conversational scenarios with per-character audio control~\cite{kong2025multitalk,wei2025mocha,wang2025interacthuman}, further increasing the applicability and realism of avatar synthesis. Most recently, MLLM-based multimodal planning enables fine-grained expression, vivid emotions, reasoned actions, and interaction with the environment through shot-level guidance for long-duration generation~\cite{jiang2025omnihuman1.5,ding2025kling-avatar}.

\section{Method}
\label{sec:method}

\begin{figure}[t]
    \centering
    \includegraphics[width=\textwidth]{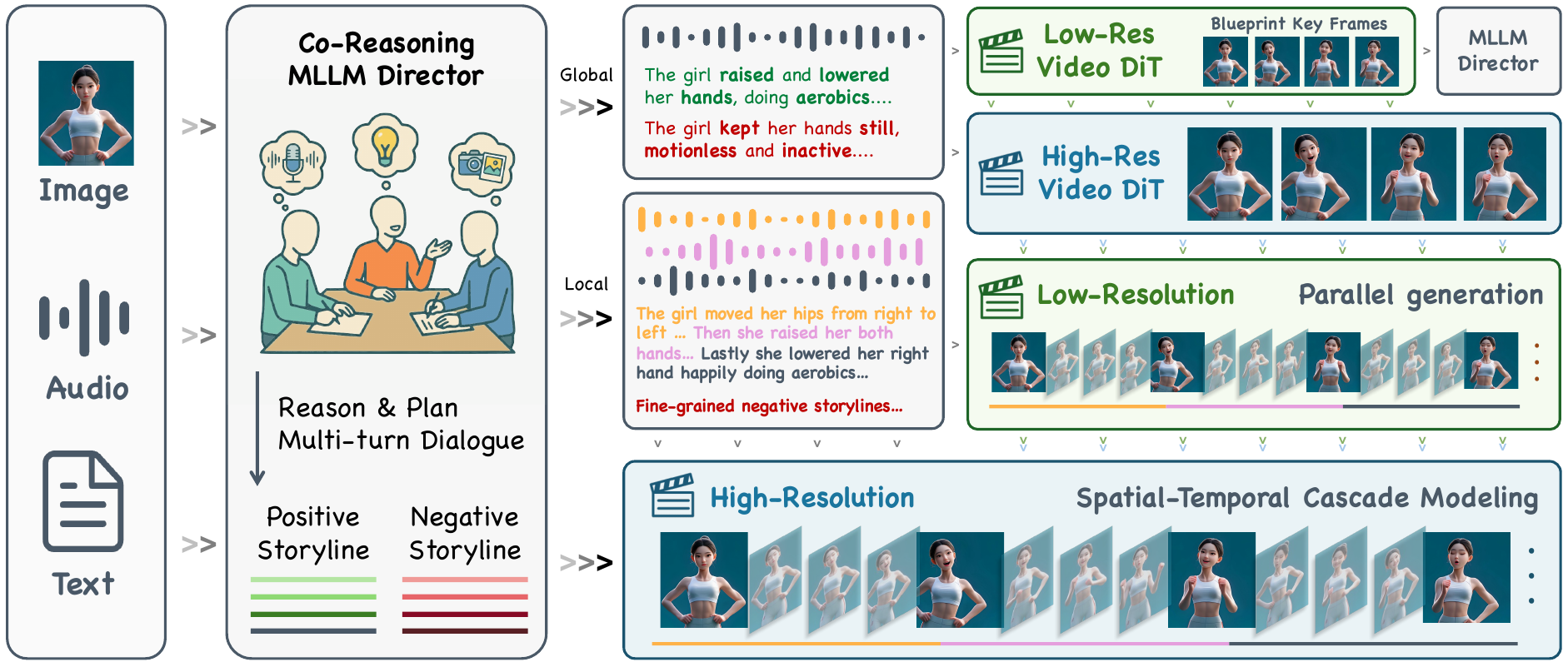}
    \caption{Overview of the KlingAvatar~2.0 framework. Given multimodal instructions, the Co-Reasoning Director reasons and plans hierarchical, fine-grained positive and negative storylines in a multi-turn dialogue manner, and the spatio-temporal cascade pipeline generates coherent, long-form, high-resolution avatar videos in parallel.}
    \label{fig:framework}
\end{figure}

KlingAvatar 2.0 extends the Kling-Avatar~\cite{ding2025kling-avatar} pipeline, as illustrated in Fig.~\ref{fig:framework}, given the reference image, input audios, and textual instructions, the system efficiently generates high-fidelity, long-form digital human videos with accurate lip synchronization and fine-grained control over multiple speakers and roles. In the following, we detail the spatial-temporal cascade diffusion framework (Sec.~\ref{subsec:cascade}), the co-reasoning multimodal storyline director (Sec.~\ref{subsec:director}), the multi-character control module (Sec.~\ref{subsec:multichar}), and the acceleration techniques (Sec.~\ref{subsec:acceleration}).

\subsection{Spatial-Temporal Cascade Modeling}
\label{subsec:cascade}

To support long-duration, high-resolution avatar synthesis with efficient computation, KlingAvatar~2.0 adopts a spatial-temporal cascade of audio-driven DiTs built on top of pretrained video diffusion models, as illustrated in Fig.~\ref{fig:framework}. The pipeline comprises two nested cascades that jointly handle global storyline planning over long horizons and local spatio-temporal refinement. First, a low-resolution diffusion model generates a blueprint video that captures global dynamics, content, and layout; representative low-resolution keyframes are then upscaled by a high-resolution DiT, enriching fine details while preserving identity and scene composition under the same Co-Reasoning Director's global prompts. Next, a low-resolution video diffusion model expands these high-resolution anchor keyframes into audio-synchronized sub-clips via first-last-frame conditioned generation, where the prompts are augmented with the blueprint keyframes to refine fine-grained motion and expression. An audio-aware interpolation strategy synthesizes transition frames to enhance temporal connectivity, lip synchronization, and spatial consistency. Finally, a high-resolution video diffusion model performs super-resolution on the low-resolution sub-clips, producing high-fidelity, temporally coherent video segments.

\subsection{Co-Reasoning Director}
\label{subsec:director}

KlingAvatar~2.0 employs a Co-Reasoning Director that jointly reasons over audio, images, and text in a multi-turn dialogue manner, building on recent MLLM-based avatar planners~\cite{jiang2025omnihuman1.5,ding2025kling-avatar}. The Director is instantiated with three experts: (i) an audio-centric expert performs transcription and paralinguistic analysis (emotion, prosody, speaking intent); (ii) a visual expert summarizes appearance, layout, and scene context from reference images; and (iii) a textual expert interprets user instructions, incorporates conversational history from the other experts, and synthesizes a logically coherent storyline plan.
These experts engage in several rounds of co-reasoning with chain-of-thought, exposing intermediate thoughts to resolve conflicts (e.g., an angry vocal tone paired with a neutral script) and to fill in underspecified details such as implied actions or camera movements. The director outputs a structured storyline that decomposes the video into a sequence of shots. Additionally, we also introduce a negative director, where positive prompts emphasize desired visual and behavioral attributes, and negative prompts explicitly down-weight implausible poses, artifacts, and fine-grained opposite emotions (e.g., sad vs.\ happy) or motion styles (e.g., overly fast vs.\ slow).

For long videos, the director further refines the global storyline into segment-level plans aligned with the audio timeline, which directly parameterize the keyframe cascade and clip-level refinement modules. This high-level multimodal planning converts loosely specified instructions into a coherent script that can be consistently followed by the diffusion backbone, substantially improving semantic alignment and temporal coherence.

\subsection{Multi-Character Control}
\label{subsec:multichar}

KlingAvatar~2.0 generalizes the single-speaker avatar setting to multi-character scenes and identity-specific audio control. Our design follows the character-aware audio-injection paradigm used in recent multi-person conversational avatars~\cite{wang2025interacthuman,kong2025multitalk,wei2025mocha}. Empirically, we observe an important architectural property: hidden features at different depths of the DiT blocks exhibit distinct feature representations. In particular, latent representations in deep DiT layers are organize into semantically coherent spatial regions with reduced noise, and these regions align well with individual characters and other salient objects.

\begin{figure}[t]
    \centering
    \includegraphics[width=\textwidth]{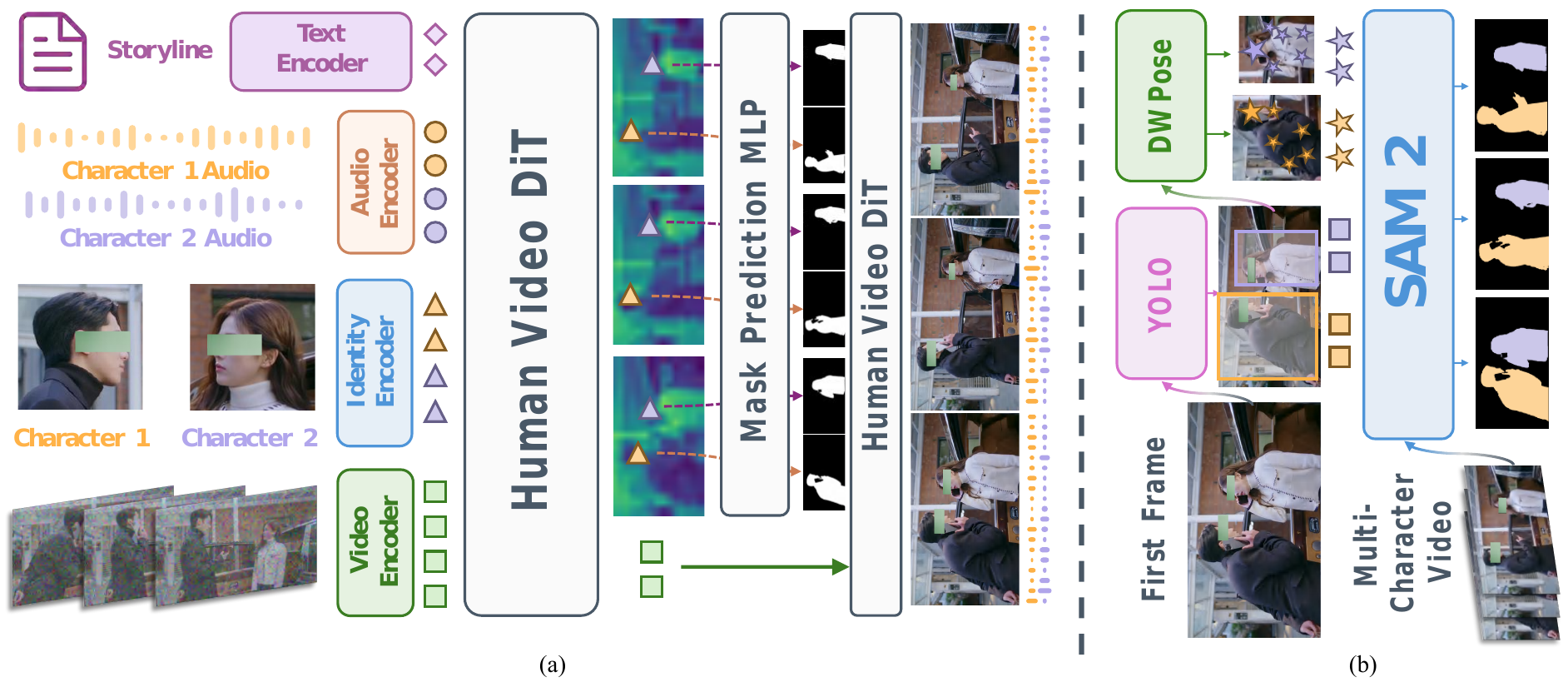}
    \vspace{-0.6cm}
    \caption{(a) Multi-character video generation pipeline with identity-specific audio control. A mask-prediction head is attached to deep DiT features, and the predicted masks gate ID-specific audio injection into corresponding regions. (b) Automated multi-character video annotation pipeline.}
    \label{fig:multichar}
\end{figure}

Motivated by this observation, we attach a mask-prediction head to selected deep DiT blocks, as shown in Fig.~\ref{fig:multichar}(a). Concretely, given a specified character in the first frame, we encode the reference identity crops using the same patchification scheme without adding noise to reference tokens. We then compute cross-attention between deep video latent tokens and these reference tokens for each identity, and apply MLP modules to regress per-frame character masks. Ground-truth (GT) masks are downsampled to match the spatial and temporal resolution of the intermediate latent features. During training, the DiT video backbone is frozen and only the mask-prediction modules are optimized. During denoising, the predicted masks are used to gate the identity-specific audio stream injection to corresponding regions.

To facilitate curation of a large-scale multi-character dataset for training, we expand our data sources to include podcasts, interviews, multi-character television serires and more.To collect GT character masks at scale, we developed an automated annotation pipeline that produces per-character video masks, as illustrated in Fig.~\ref{fig:multichar}(b). The pipeline leverages serveral expert models: YOLO~\cite{yolo2023} for person detection, DWPose~\cite{yang2023effective} for keypoint estimation, and SAM2~\cite{ravi2024sam2} for segmentation and temporal tracking. Specifically, we detect all characters in the first frame with YOLO, estimate keypoints within each detection using DWPose, and use the resulting bounding boxes and keypoints as prompts for SAM2 to segment and track each person in subsequent frames. Finally, we validate the generated video masks against per-frame YOLO and DWPose estimation results and filter out misaligned or low-overlap segments to ensure high-quality annotations for training.

\subsection{Accelerated Video Generation}
\label{subsec:acceleration}

To achieve accelerated inference efficiency, we explored distillation schemes based on trajectory-preserving distillation exemplified by PCM~\cite{wang2024phased} and DCM~\cite{lv2025dualexpertconsistencymodelefficient}, and distribution matching distillation exemplified by DMD ~\cite{yin2024onestepdiffusiondistributionmatching}. Based on comprehensive evaluations of experimental cost, training stability, inference flexibility, and final generative performance metrics, we ultimately selected the trajectory-preserving distillation approach. To further enhance distillation efficiency, we developed customized time schedulers by analyzing the performance of the base model across different timesteps, thereby balancing the inference speedup ratio against model performance. Within our distillation algorithm, we introduced a multi-task distillation paradigm through a series of precisely designed configurations.
This paradigm not yields a synergistic effect (1+1>2), improving the distillation outcomes for each individual task.

\begin{table}[t]
\centering
\caption{Quantitative results of GSB metrics between our approach and other competitors across diverse criteria.}
  \vspace{-0.1cm}
  \label{tab:gsb}
  \resizebox{\textwidth}{!}{
    \begin{tabular}{l|ccccccc}
    \toprule
        GSB & Overall & Face-Lip Sync. & Visual Qual. & Motion Qual. & Motion Expr. & Text Rel. \\
    \midrule
    Ours vs. HeyGen & 1.26 & 0.86 & 1.76 & 0.88 & 1.53 & 1.39 \\
    Ours vs. Kling-Avatar & 1.73 & 0.80 & 0.89 & 1.13 & 2.47 &  3.73 \\
    Ours vs. OmniHuman-1.5 & 1.94 & 1.02 & 1.99 & 1.06 & 1.13 & 1.08 \\
    \bottomrule
    \end{tabular}
}
\end{table}
\begin{figure}[t]
    \centering
    \vspace{-0.2cm}
    \includegraphics[width=\linewidth]{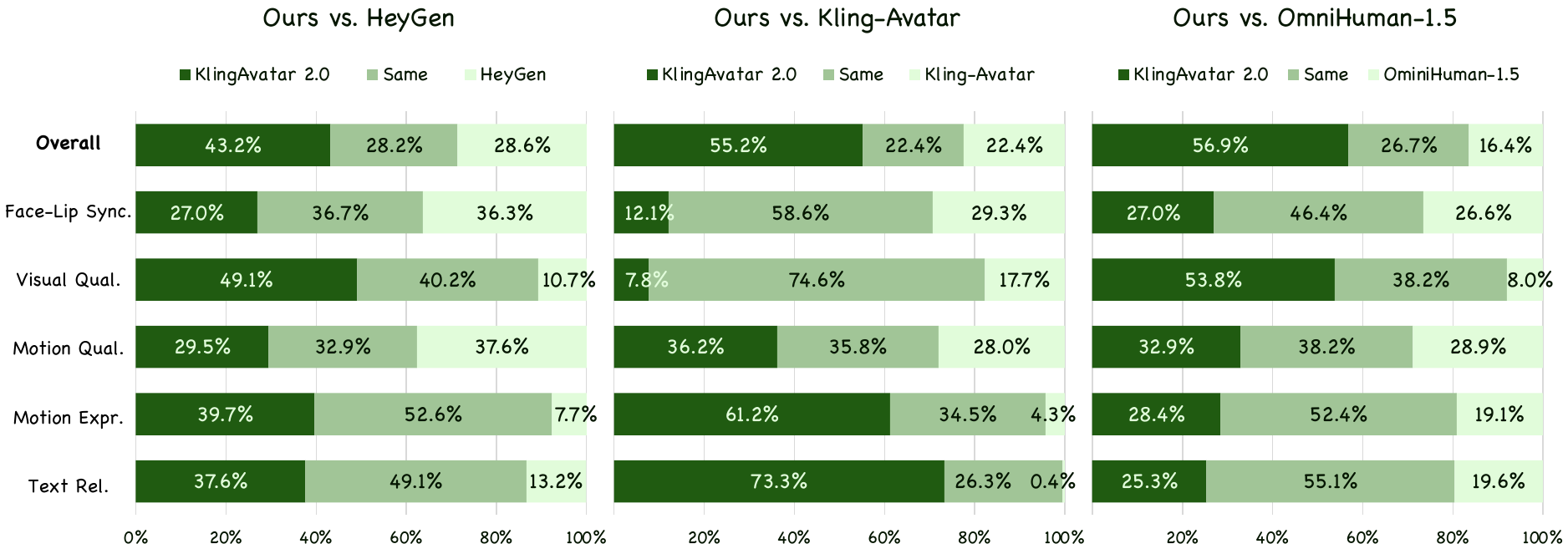}
    \vspace{-0.6cm}
    \caption{Visualization of GSB benchmark results comparing KlingAvatar~2.0 with HeyGen, Kling-Avatar, and OmniHuman-1.5 across various evaluation criteria.}
    \label{fig:quantitative}
\end{figure}

\begin{figure}[t]
    \centering
    \includegraphics[width=\linewidth]{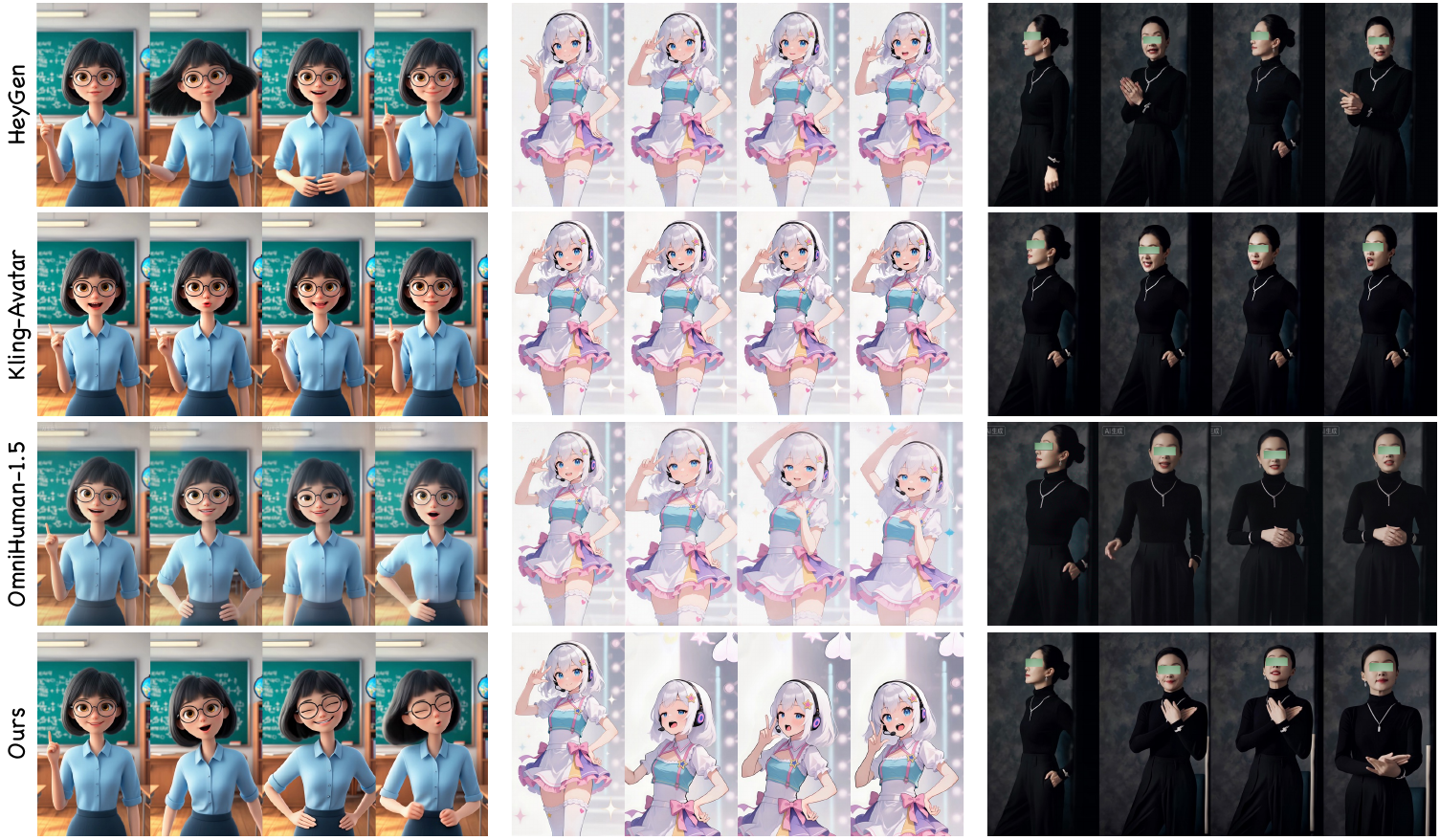}
    \caption{Qualitative comparison between KlingAvatar~2.0 and baseline methods. \textbf{Left}: Our method produces more natural hair dynamics and vivid facial expressions. \textbf{Middle}: Our results adhere more closely to the specified bottom-to-top camera motion. \textbf{Right}: Our generated video aligns better with the prompt ``...turned to the front and folded her hands in front of her chest''.}
    \label{fig:baseline-comparison}
\end{figure}

\section{Experiments}
\label{sec:exp}

\subsection{Experimental Setup}

We follow the human preference–based subjective evaluation protocol in \cite{ding2025kling-avatar} to conduct comprehensive evaluations of KlingAvatar~2.0. We construct 300 high-quality test cases, each consisting of paired image, audio, and text prompts, including 100 Chinese speech, 100 English speech, and 100 singing samples. For each case, annotators perform Good/Same/Bad (GSB) pairwise comparisons between our results and those of baseline methods. We report (G+S)/(B+S) as the main metric, where higher scores indicate stronger human preference. To capture fine-grained aspects of video quality and multimodal alignment, we further extend GSB to a richer set of detailed criteria, including:

\begin{itemize}
    \item \textbf{Face–Lip Synchronization.} Measures temporal alignment between face and lip motions and speech, the naturalness and continuity of facial expressions, and the consistency between movements, facial dynamics, and phonetic content.
    \item \textbf{Visual Quality.} Evaluates overall aesthetics, sharpness, and fidelity of fine details such as hair, teeth, and skin texture, as well as temporal consistency of appearance and robustness to artifacts, flickering, and implausible lighting or color.
    \item \textbf{Motion Quality.} Assesses the plausibility, smoothness, and temporal coherence of body, head, and camera motion, avoiding geometric distortions, jitter, unnatural warping, body-part collapses, and unstable character tracking.
    \item \textbf{Motion Expressiveness.} Characterizes the richness, diversity, and vividness of lip, facial, and full-body movements, and their emotional match with the audio and text in terms of intensity, timing, and modulation of gestures and head poses.
    \item \textbf{Text Relevance.} Reviews the alignment and coherence of the generated storyline, camera trajectories, and scene dynamics with the textual instructions.
\end{itemize}

\subsection{Experimental Results}

We compare KlingAvatar~2.0 against three strong baselines: HeyGen~\cite{heygen}, Kling-Avatar~\cite{ding2025kling-avatar}, and OmniHuman-1.5~\cite{jiang2025omnihuman1.5}. Quantitative GSB results across diverse dimensions are summarized in Table~\ref{tab:gsb} and visualized in Fig.~\ref{fig:quantitative}. Our method achieves strong overall performance, with especially notable improvements in motion expressiveness and text relevance. Qualitative comparisons are provided in Fig.~\ref{fig:baseline-comparison}.

\begin{figure}
    \centering
    \includegraphics[width=\linewidth]{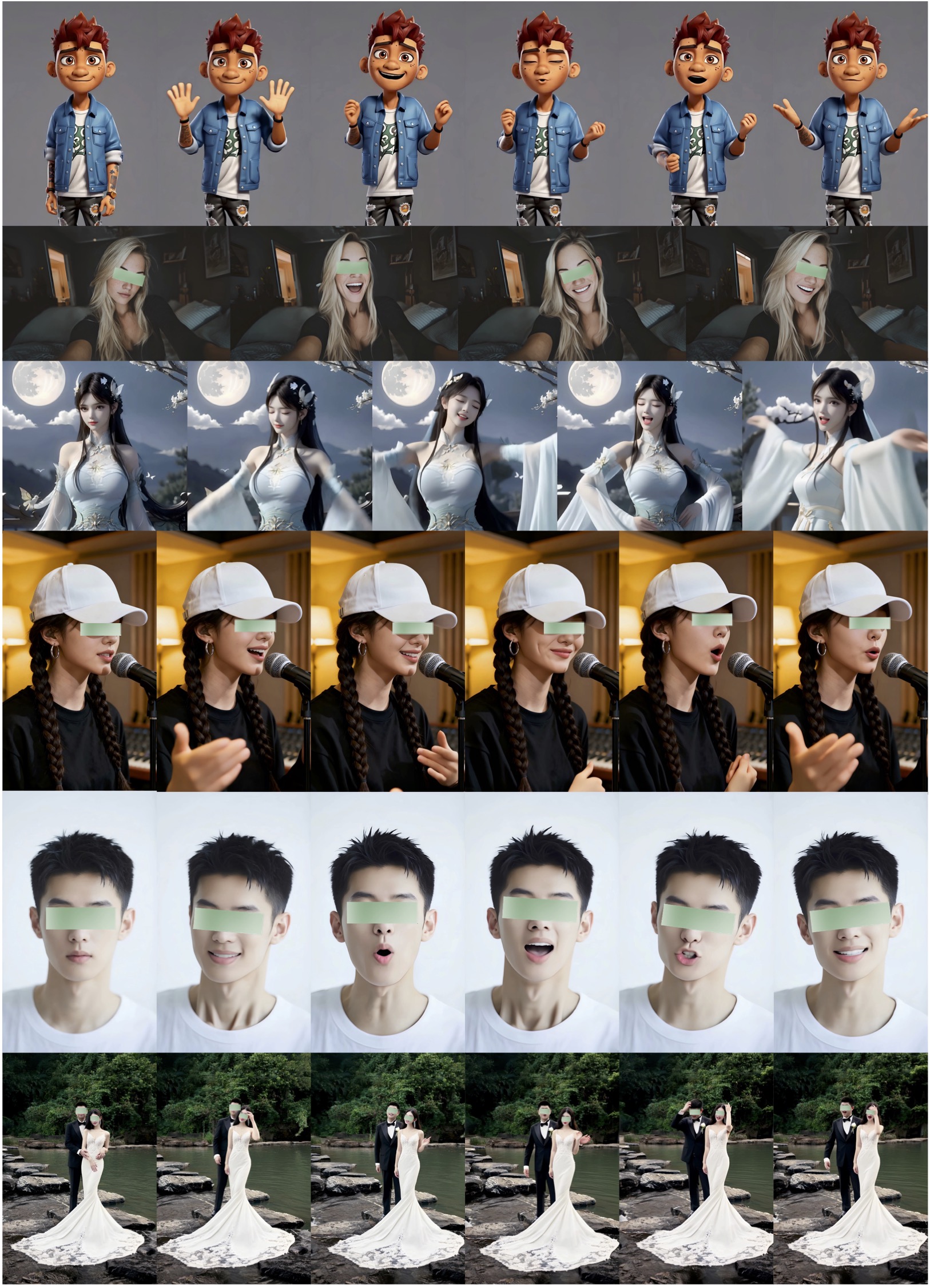}
    \caption{Representative qualitative results generated by our spatial–temporal cascade framework with the multimodal co-reasoning director.
    }
    \label{fig:qualitative}
\end{figure}

\begin{figure}[t]
    \centering
    \includegraphics[width=\linewidth]{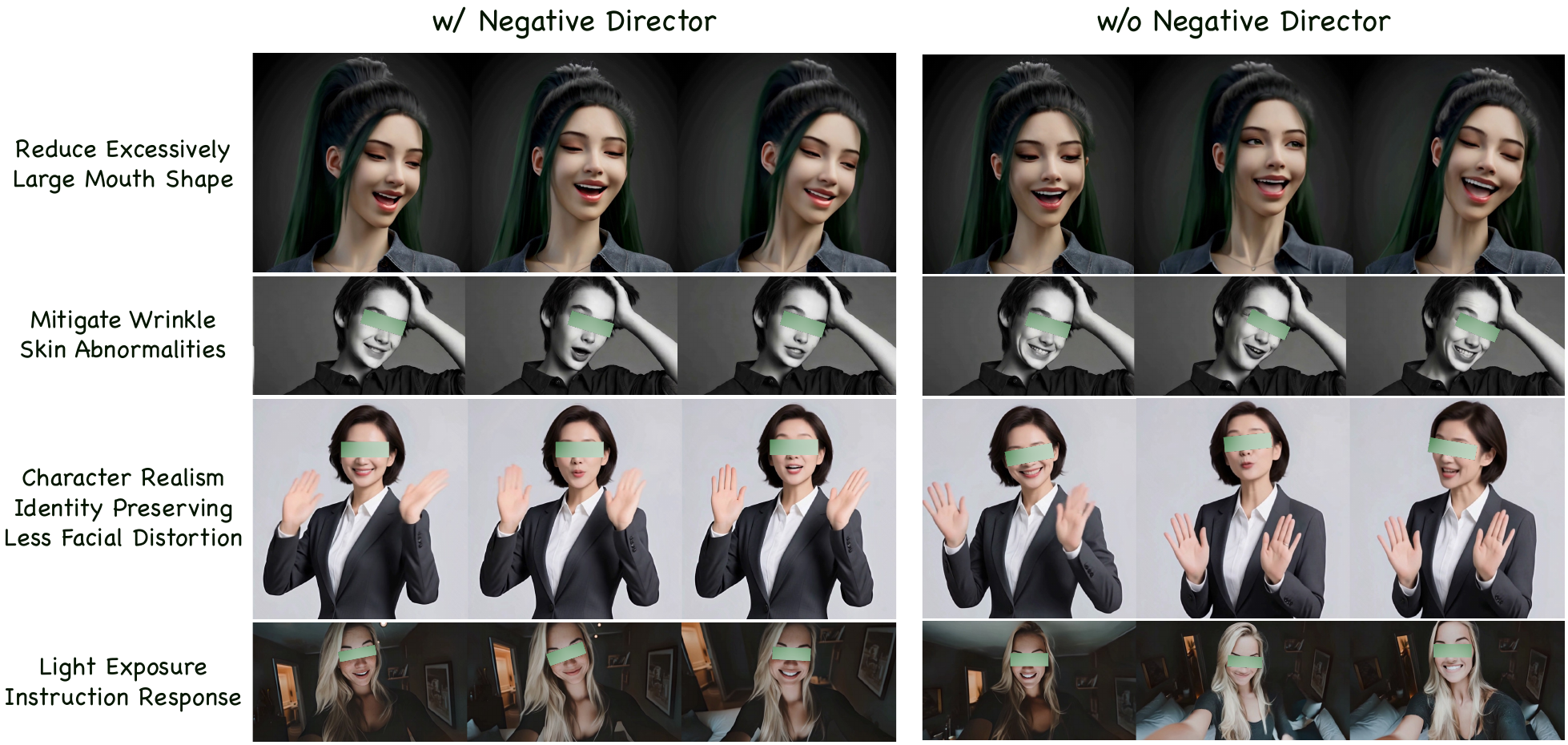}
    \caption{Ablation study of the negative director on blueprint keyframes. The negative director enhances facial expressiveness, improves temporal stability and emotional controllability, and reduces lighting and exposure artifacts.}
    \vspace{-0.2cm}
    \label{fig:negative-director}
\end{figure}

Our model generates richer, more natural dynamic effects and follows multimodal instructions more faithfully, demonstrating an enhanced understanding of complex audiovisual intent and effectively addressing the key limitations of baseline methods. Across baselines, hair dynamics are either relatively rigid (Kling-Avatar, OmniHuman-1.5) or occasionally less physically grounded (HeyGen). Our method produces more temporally consistent and physically plausible hair motion and head poses, leading to improved perceived naturalness. For multimodal instruction following, HeyGen and Kling-Avatar generally generate stable but relatively simple camera trajectories, while OmniHuman-1.5 sometimes deviates from the specified camera instructions. KlingAvatar~2.0 produces camera motions and scene dynamics that are more closely aligned with the textual prompts, yielding detailed and coherent interpretations. Regarding emotional expression and fine-grained motion instructions, HeyGen and Kling-Avatar sometimes under-emphasize the target actions, whereas OmniHuman-1.5 incorrectly folds the hands at the waist instead of in front of the chest. In contrast, our approach more reliably captures the intended motion of folding the hands in front of the chest, produces movements synchronized with the audio and target emotion, and yields facial and body expressions that are both expressive and realistic.

Fig.~\ref{fig:qualitative} showcases results generated by our framework across diverse scenarios. Powered by the spatial–temporal cascade and the multimodal co-reasoning director, our approach accurately interprets and fuses image, audio, and text instructions, producing emotionally expressive characters, coherent full-body and camera motions, and precise, fine-grained lip synchronization. Beyond single-speaker talking scenarios, our method generalizes well to multi-person interactions with per-character audio conditioning. These results demonstrate the robustness and versatility of KlingAvatar~2.0 in open, complex settings.

As shown in Fig.~\ref{fig:negative-director}, prior work typically uses a small, fixed set of generic negative prompts (e.g., ``artifacts, bad quality, blur'') for an entire video, offering only coarse control over undesired content. In contrast, our negative director employs detailed, shot-specific negative prompts that track the evolving storyline and target emotion. This per-shot control discourages implausible expressions, unstable motion, and narrative-inconsistent artifacts, leading to more natural, emotionally faithful, and temporally stable results that better follow the text description.

\section{Conclusion}

In this paper, we present KlingAvatar~2.0, a unified framework that enables spatio-temporal cascade generation for high-resolution, long-duration, lifelike multi-person avatar videos with omni-directed co-reasoning directors. Our multimodal, multi-expert co-reasoning director thinks and plans over audio cues, visual contexts, and complex instructions through multi-turn dialogues to resolve ambiguities and conflicting signals, producing coherent global storylines to guide the long-form synthesis trajectory and detailed local prompts to refine sub-clip dynamics. The hierarchical storyline drives generation of low-resolution blueprint keyframes, and spatio-temporal upscaled high-resolution, audio-synchronized sub-clips, which are efficiently composed into long-form videos in parallel via first–last frame conditioning. We further extend the application scenarios to multi-character settings with identity-specific audio control and develop an automated annotation pipeline to curate large-scale multi-person video datasets. Experiments demonstrate that KlingAvatar~2.0 delivers leading performance in visual fidelity, identity preserving, lip–audio synchronization, instruction-following, long-duration coherence, and multi-character, multi-audio controllability. We believe our exploration of an omni-directed, multi-character, multi-audio, long-form, high-resolution avatar synthesis framework paves the way for future research and applications in digital human generation.

\newpage
\section{Contributors}
\label{sec:contributors}

All contributors are listed in alphabetical order by their last names.

Jialu Chen, Yikang Ding, Zhixue Fang, Kun Gai, Yuan Gao, Kang He, Jingyun Hua, Boyuan Jiang, Mingming Lao, Xiaohan Li, Hui Liu, Jiwen Liu, Xiaoqiang Liu\footnote{Project Lead}, Yuan Liu, Shun Lu, Yongsen Mao, Yingchao Shao, Huafeng Shi, Xiaoyu Shi, Peiqin Sun, Songlin Tang, Pengfei Wan, Chao Wang, Xuebo Wang, Haoxian Zhang, Yuanxing Zhang, Yan Zhou.


\clearpage
\newpage
\bibliographystyle{kling/plainnat}
\bibliography{paper}

\clearpage
\newpage

\end{document}